\documentclass{article}
\usepackage{spconf,amsmath,graphicx}
\usepackage{subcaption}
\usepackage{color}

\title{Image-level Classification in Hyperspectral Images using Feature Descriptors, with Application to Face Recognition}
 \twoauthors
  {Vivek Sharma$^{\diamond}$}
{$^{\diamond}$ ESAT-PSI, VISICS, iMinds, KU Leuven\\
        {\tt\small \{vivek.sharma,luc.vangool\}@esat.kuleuven.be}}%
  {Luc Van Gool$^{\diamond \mp}$\thanks{This work is supported by the European Commission's Seventh Framework  Programme as part of the project ROVINA.}}
{$^{ \mp}$  Computer Vision Laboratory, ETH Z\"{u}rich \\
        {\tt\small vangool@vision.ee.ethz.ch}}%

\begin{document}
%
\maketitle
\begin{abstract}
Image-level classification from hyperspectral images (HSI) has seldom been addressed in the literature. Instead, traditional classification methods have focused on individual pixels. Pixel-level HSI classification comes at a high computational burden though. In this paper, we present a novel pipeline for classification at image-level, where each band in the HSI is considered as a separate image. In contrast to operating at the pixel level, this approach allows us to exploit higher-level information like shapes. We use traditional feature descriptors, i.e. histograms of oriented gradients, local binary patterns, and the scale-invariant feature transform. For demonstration we choose a face recognition task. The system is tested on two  hyperspectral face datasets, and our experiments show that the proposed method outperforms the existing state-of-the-art hyperspectral face recognition methods.

\end{abstract}

\section{Introduction}
\begin{figure}[t] 
\centering
{\includegraphics[width=0.50\columnwidth]{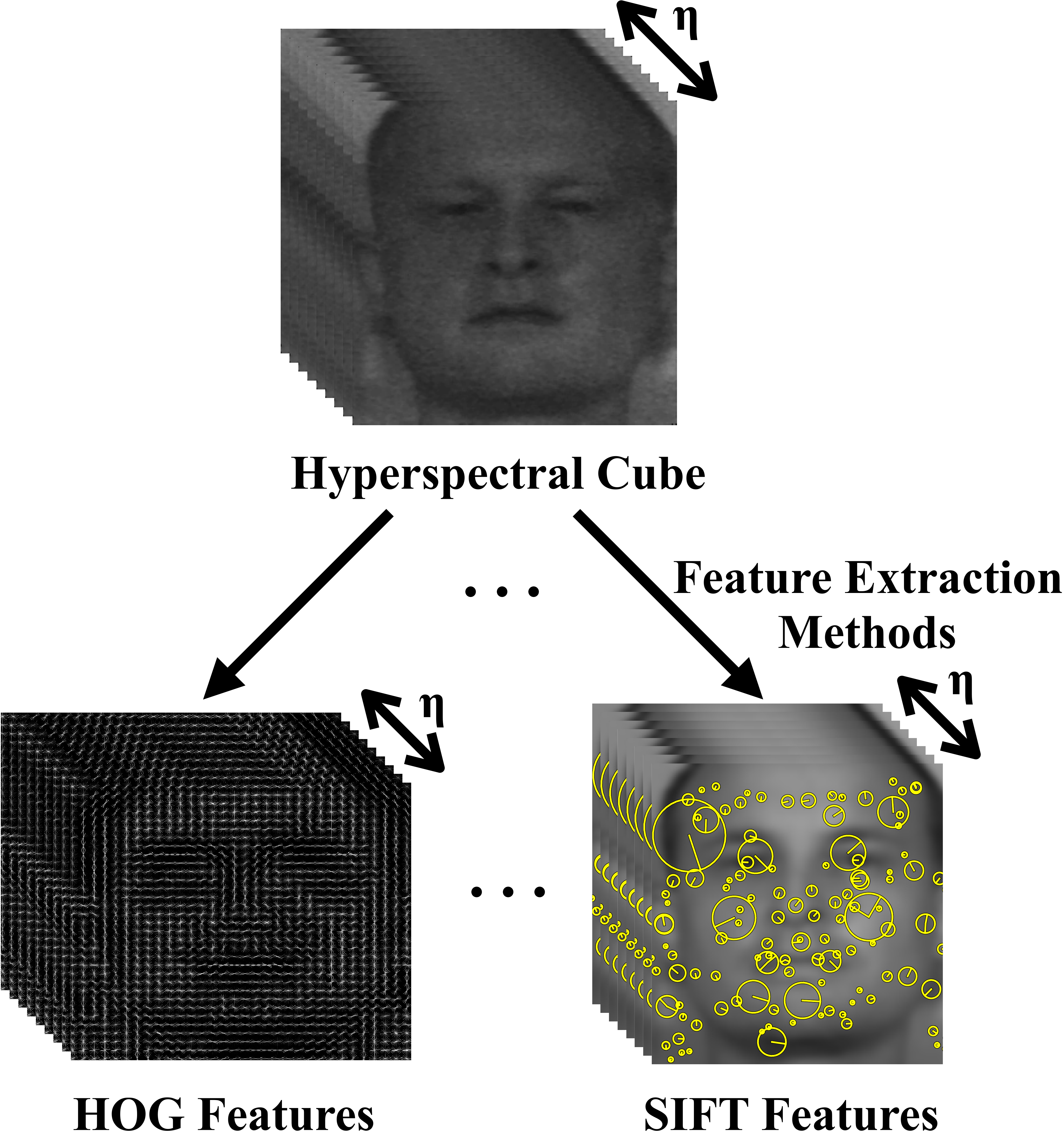}
} 
\caption{For a given hyperspectral image, with $\eta$ the number of spectral bands, several features are extracted from each of the $\eta$ mono-spectral images.} \label{fig:frontpage}
\end{figure}

A hyperspectral image (aka. hyperspectral cube) consists of two spatial dimensions and a spectral dimension. The latter contains information pertinent to the intrinsic material properties of the object. This spectral information in HSI makes them well suited for classification tasks, such as  scene recognition~\cite{brown2011multi}, 3D reconstruction~\cite{zia20153d}, saliency detection~\cite{liang2013salient}, pedestrian detection~\cite{hwang2015multispectral,luo2010pedestrian}, material classification~\cite{salamati2009material}, cultural heritage~\cite{balas2003novel} and many more. In this paper, we propose a  framework for hyperspectral image classification, where each band in the HSI is considered as a separate image. Treating the problem at the image-level allows us to exploit high-level information, like shapes, that help to improve the classification performance.  We use traditional feature descriptors (i.e. SIFT~\cite{sift}, HOG~\cite{hog}, LBP~\cite{lbp}) for image-level feature extraction and classification (See Fig.~\ref{fig:frontpage} for illustration). For the demonstration of our method, we investigate into
face recognition using wavelengths ranging from the visible ($380-700nm$) to the near-infrared ($750-1100nm$) spectrum.

\begin{figure*}[t] 
\centering
{\includegraphics[width=1.75\columnwidth]{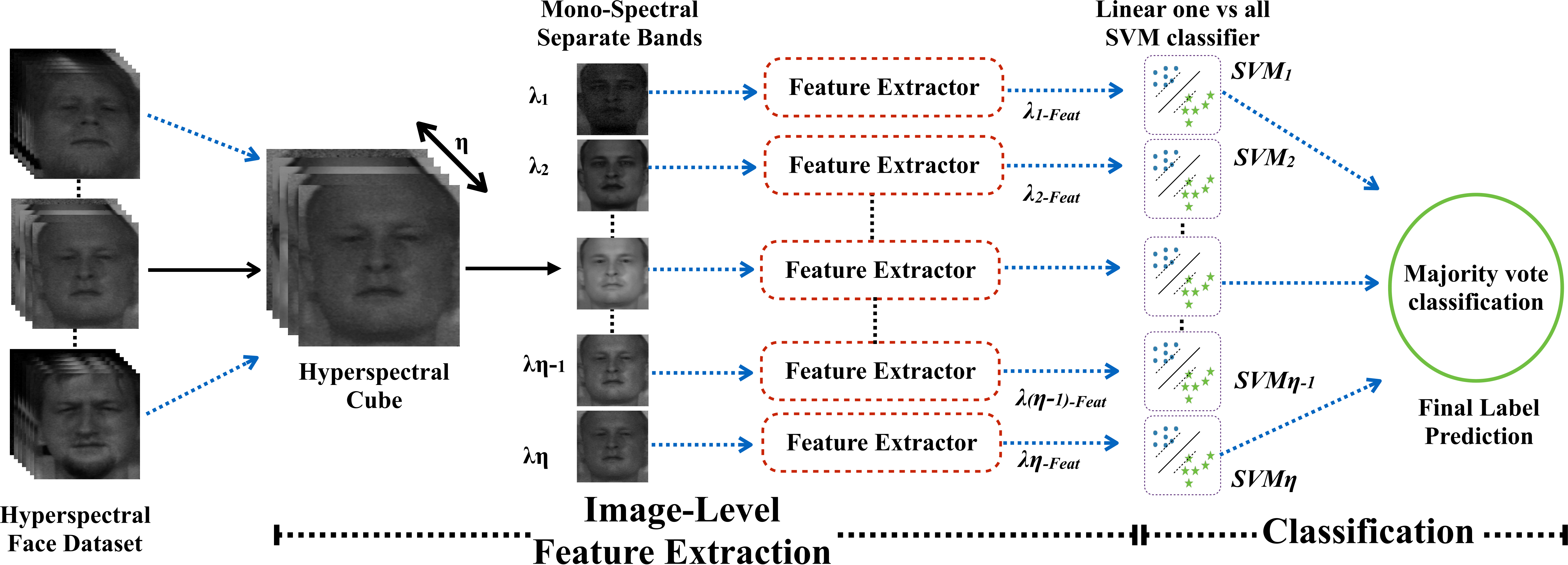}
} 
\caption{An overview of our image-level classification chain in the hyperspectral  images. Each band in the HSI is treated as a mono-spectral image (or band).  Features ($Feat$) are extracted for each image using a feature extraction method (e.g. HOG, LBP, and SIFT), and then these features are fed to linear one vs all SVM classifier, where $\eta$ SMVs  are trained on $\eta$ bands. For the final label prediction, we merge the predictions of all SVM learners trained on different bands using majority voting.} \label{fig:scnn}
\end{figure*}

The availability of more bands than the usual three RGB bands has been shown advantageous in disambiguating objects.  In literature, the common tendency to exploit the hyperspectral data has been addressed at pixel-level.  As the spatial and spectral dimensions in HSI increase, it is difficult to separate hyperspectral data at the pixel-level for large classes dataset using statistical methods~\cite{stat1,stat2}. However, in our method classification is done at the image-level, where from each image we extracted a limited number of features, far fewer than the number of pixels. With such smaller numbers of  features at hand, we can afford to not further reduce the number of spectral bands that are considered. This allows us to exploit the entire input  space without losing information for feature extraction. We show that the entire feature space information leads to significantly improved performance in a face recognition task. 

Recently, deep learning (DL) a learning-based representation has outperformed traditional descriptors for deriving distinctive features in image classification task~\cite{deepface,vgg}, but DL has a major shortcoming: it requires many samples for the training process and an insufficient number of training  samples quickly leads the network to overfitting.  As we already know, in HSI the spectral bands are very unique and discriminative.  Thus,  we believe traditional feature descriptors can effectively exploit this spectral information  and help to generate powerful feature representation of their content that characterize the object better.  In this work, we have explored the hand-crafted HSI features in V-NIR images and also have shown improvement in the classification performance by a significant margin for a face recognition task.

The rest of the paper is organized as follows. In Section~\ref{sec:relatedwork}, we discuss the related work, and  Section~\ref{sec:method} describes our proposed method. Experiments and analysis are given in Section~\ref{sec:experiments}, and Finally, the conclusions are drawn in Section~\ref{sec:conclusion}.

\section{Related Work}\label{sec:relatedwork}

There is an extensive work on HSI classification, but here we mention a few relevant papers on face recognition task only. During the last decade, face recognition has achieved great success.  The research has tended to focus on RGB or b/w images, rather less attention has been paid to hyperspectral images. In this paper, we focus on HSI face recognition. The idea of face recognition in hyperspectral images, started with Pan et al.~\cite{spectralsignature}, who manually extracted the mean spectral signatures from the human face in the NIR spectrum, and were then compared using Mahalabonis distance.   Further in~\cite{spectraleigenface} Pan et al.  extended their work~\cite{spectralsignature} by incorporating spatial information in addition to spectral. Similarly, Robila et al.~\cite{spectralangle} also uses spectral signatures but their comparison criterion was spectral angle measurements. Almost all the existing proposed HSI face recognition methods perform dimensionality reduction and the low-dimensional feature space are extracted for classification. Di et al.~\cite{pca} projected the hyperspectral image into low dimensional space using 2D-PCA for feature extraction, and were then compared using Euclidean distance.  In recent works, Shen and Zheng~\cite{gabor} apply 3D Gabor wavelets with different central frequencies to extract signal variances in the hyperspectral data. Liang et al.~\cite{ldp} utilize 3D high-order texture pattern descriptor to extract distinctive micro-patterns which integrate the spatial-spectral information as features. Uzair et al.~\cite{dct} apply 3D discrete cosine transformation and extract the low frequency coefficients as features, where Partial Least Squares Regression is used for classification. Uzair et al. most recent work in~\cite{pls} extended their previous study~\cite{dct} and employed 3D cubelets to extract the spatio-spectral covariance features for face recognition.  In contrast, in our work, the hyperspectral image is fully exploited without losing information by performing feature extraction and  classification at the image-level.

\begin{figure*}[htb] \label{fig:tsne}
    \centering
    \begin{subfigure}[b]{0.45\textwidth}
        \includegraphics[width=0.99\textwidth]{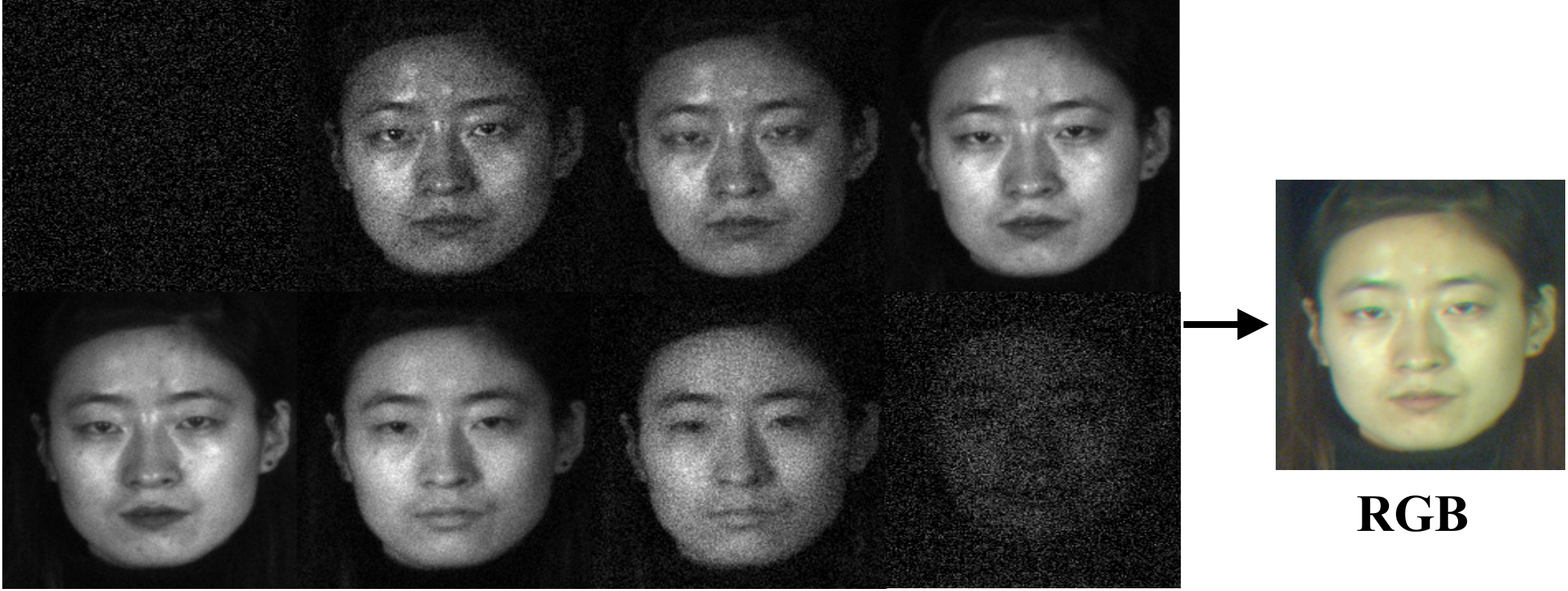}
        \caption{\textbf{PolyU-HSFD: Visible Range (440-690nm). Bands with a  step size of  40nm are shown.}}
        
    \end{subfigure}\hspace{1cm}
    ~ 
    \begin{subfigure}[b]{0.45\textwidth}
    
        \includegraphics[width=0.83\textwidth]{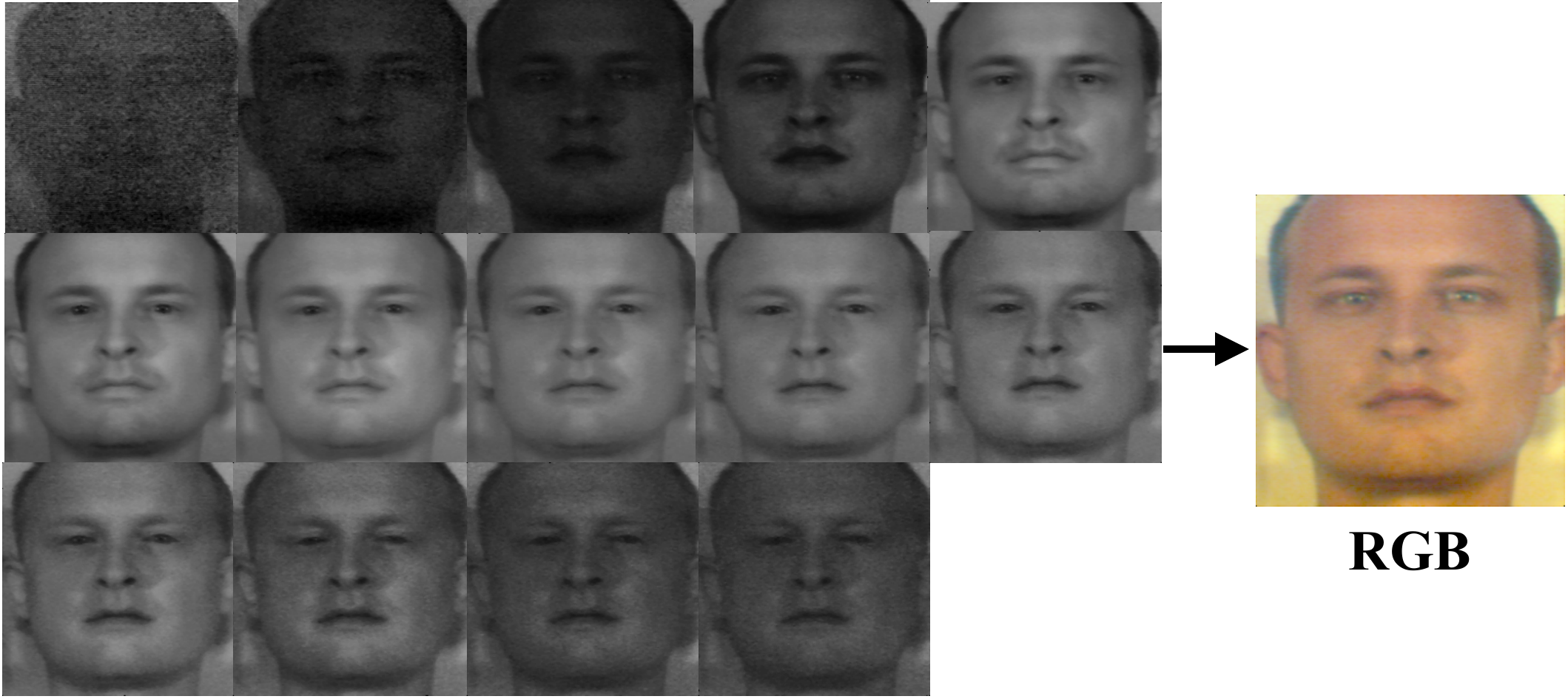}
        \caption{\textbf{CMU-HSFD: V-NIR Range (450-1090nm). Bands with a  step size of  50nm are shown.}}

    \end{subfigure} 
    \caption{Example hyperspectral image for a subject with its corresponding RGB image.}\label{fig:ex_dataset}
\end{figure*}

\section{Proposed Method} \label{sec:method}

In the last two decades, researchers have developed robust descriptors (feature extraction methods) to extract useful feature representations from the image that are highly distinctive and are perfect for an efficient and robust object recognition. The traditional feature descriptors (SIFT, HOG, LBP and more) have several advantages. These features are  invariant to image scaling,  geometric and photometric transformations: translations or rotations,  common object variations, minimally affected by noise and small distortions, and to a great extent invariant to change in illumination.  
These descriptors when combined with hyperspectral images,  make the extracted features even more robust and powerful. We believe that the extracted new class of features obtained from hyperspectral images shall give near-perfect separation because these features are generated from highly discriminative bands (Fig.~\ref{fig:ex_dataset}) in HSI that are captured in different wavelengths.  These images captured in different wavelength range, contain discriminative information that characterize the object better with great precision and detail.

Fig.~\ref{fig:scnn} shows the schematic layout of our framework for hyperspectral image classification. In our work, each mono-spectral band in the HSI is treated as a separate image. Treating the mono-spectral bands as separate images, allows us to exploit: the discriminative texture patterns present in the images  captured in different wavelengths, hence able to utilize the entire space without losing information for feature extraction. Then, the features are extracted for each mono-spectral band using a feature extraction method (e.g. HOG, LBP, and SIFT). This allows us to capture the high-level information like shapes and abstract concepts from images: making it more suitable for high-level visual recognition tasks, similar is not possible at pixel-level. Classification at the pixel-level (think of raw pixel values) comes at high computational burden, and it turns out that, it is difficult to disambiguate objects with large classes dataset by simple concatenating of multi-spectral data into single pixel-related vectors using statistical methods. Thus, we extract from each image a limited number of features, far fewer than the number of pixels. With such smaller numbers of  features at hand, we show that, we achieve better recognition results, significantly outperforming pixel-level features (see Table~\ref{table:soa}), and can afford to not further reduce the number of spectral bands that are considered. Then, the extracted feature vectors are fed to linear one vs all SVM classifier, where we have a SVM for each band. For the final label prediction, we merge the predictions of all SVM learners trained on different bands using majority voting. 

For feature extraction of each mono-spectral image, we use a fixed-size representation to extract SIFT, LBP and HOG features.  The parameters for feature extraction (i.e. window size,  step size, number of oriented bins, number of clusters, and more) contribute to the good results, and play a crucial role in improving the recognition performance. In Section~\ref{sec:experiments}, we describe the implementation setup in detail. 

The performance of the the proposed approach is demonstrated for face recognition in HSI.  Excellent results show that these robust feature descriptors are perfect for reliable face recognition in hyperspectral images.

\section{Experiments} \label{sec:experiments}

The proposed method in the previous section was tested on two standard hyperspectral face datasets~\cite{pca,cmu} for a face recognition task. Our experiments consist of two parts. In the first part,
 we compare our proposed method with the state-of-the-art HSI face recognition methods. Then in the second part, we compare the usefulness of the  discriminative spectral bands with the RGB image representation of HSI.  We  now introduce the datasets used and then we move onto  training/testing protocol, and implementation details.

\subsection{Experimental Setup}

For the experimental evaluation, we used a desktop with Intel i7-2600K CPU at 3.40GHZ, 8GB RAM. All experiments were performed using vlfeat library~\cite{vlfeat}.

\paragraph*{\textbf{Hyperspectral Face Datasets:}} 

Hong-Kong Polytechnic University Hyperspectral Face Dataset (PolyU-HSFD)~\cite{pca} (see Fig.~\ref{fig:ex_dataset}  and Table~\ref{table:dataset}) is acquired using the CRI's VariSpec LCTF with a halogen light source. The database contains data of 48 subjects acquired in multiple sessions, with 1-7 cubes over all sessions. Following the same experimental protocol of~\cite{ldp,dct,pls}, only first 25 subjects were used for evaluation. The acquired images for the first 6 and the last 3 bands are noisy (i.e. very high shot noise). So they are discarded from the experiment, as suggested in the previous work. In all, 24 spectral bands were used with spectral interval (i.e. step size) of 10nm.  The database has significant appearance variations of the subjects because it was constructed over a long period of  time. Major changes in appearance variation were in hair-style and skin condition.

Carnegie Mellon University  Hyperspectral Face Dataset (CMU-HSFD)~\cite{cmu}  (see Fig.~\ref{fig:ex_dataset}   and Table~\ref{table:dataset})  is acquired using the CMU developed AOTF with three halogen light sources. The database contains data of 54 subjects acquired in multiple sessions, with 1-5 cubes over all sessions. Following the same experimental protocol of~\cite{ldp,dct,pls}, only first 48 subjects were used for evaluation. Each subject maintains a still pose with negligible head movements and eye blinks, due to of that a slight misalignment error exists between individual bands because the image capturing process takes several seconds.

\begin{table}[t]
\begin{center}
\resizebox{6.9cm}{!} {
\begin{tabular}{|r|l l l l| }
  \hline
Dataset 		&   Spectral Range &  Bands & Spatial Range & Step Size  \\
\hline
PolyU-HSFD 		& 400-720nm  & 33 &$220\times180$ & 10nm\\
CMU-HSFD   		& 450-1090nm & 65 &$640\times480$ & 10nm\\
\hline
\end{tabular}}\vspace{-0.5cm}
\end{center}\caption{Hyperspectral Face Dataset}
\label{table:dataset}
\end{table}

\paragraph*{\textbf{Training and Testing Protocol:}} 
Following the same training and testing protocol as defined in~\cite{ldp,dct,pls}, only frontal view has been considered  for evaluation of both datasets. When only one sample per subject is used for training and two samples per subject are used for testing. Both gallery set (or training set) and probe set (or testing set)  were constructed for 5 times by random selection and the average recognition accuracy was reported. 

\paragraph*{\textbf{Data Preprocessing:}}  All images were cropped and resized to size  $263\times263$ in spatial dimension. Due to of low signal-to-noise ratio (i.e. high shot noise) in the datasets, we apply a median filter of size $3\times3$ to remove shot noise. 

\paragraph*{\textbf{Implementation Details:}} To extract the HOG and LBP features,  we use a window size of
 $8\times8$, and number of orientation bins of 9 for HOG.  For SIFT, we use a bin size of 4, step size of 8, then the extracted SIFT-features are Fisher encoded. To compute Fisher encoding, we build a visual dictionary using GMM with 100 clusters. We normalize the features using L2-normalization. We denote dense SIFT Fisher vectors by DSIFT-FVs. These parameters are fixed for all descriptors. We use linear SVM with $C=10$ from LIBSVM~\cite{libsvm} to train classifier with features.

\subsection{Results}  

\paragraph*{\textbf{Comparison with state-of-the-art methods:}}

In Table~\ref{table:soa}, we  quantitatively evaluate the recognition accuracy of our proposed method, and compare it with state-of-the-art hyperspectral  face recognition methods reported in the literature.  We observe from the results that among the traditional feature descriptors,  as expected DSIFT-FVs  outperforms the LBP and HOG  recognition accuracy by a significant margin. Also, it should be noted that DSIFT-FVs outperforms all the traditional methods
listed in the literature and achieve state-of-the-art accuracy with 96.1\% on PolyU-HSFD dataset.  Though, DSIFT-FVs is inferior to Band fusion+PLS~\cite{pls}, but is still better than all the other methods on CMU-HSFD dataset. Furthermore, this examination reveals that  SIFT shows the same trend of performing better in face recognition in hyperspectral images as in RGB images.

\begin{table}[t]
\begin{center}
\resizebox{6cm}{!} {
\begin{tabular}{|r|l l | }
\hline
Methods  					&   Accuracy &   \\
\hline 
 						& PolyU-HSFD & CMU-HSFD      \\
\hline
 Spectral Signature~\cite{spectralsignature} 			& 24.6	& 38.1 \\
Spectral Angle~\cite{spectralangle}			& 25.4 	& 38.1 \\
Spectral Eigenface~\cite{spectraleigenface}			& 70.3 	& 84.5 \\
2D PCA~\cite{pca}			& 71.1 	& 72.1 \\ 
3D DCT~\cite{dct}			& 84.0 	& 88.6 \\ 
LBP~(\textbf{ours})~\cite{lbp}		& 85.6  & 86.1 \\
3D Gabor Wavelets~\cite{gabor}		& 90.1 	& 91.6 \\
HOG~(\textbf{ours})~\cite{hog}		& 92.3	& 91.5 \\ 
3D LDP~\cite{ldp}		& 95.3	& 94.8 \\ 
\hline
Band fusion+PLS~\cite{pls}		& 95.2  & \textbf{99.1} \\
DSIFT-FVs~(\textbf{ours})~\cite{sift}		& \textbf{96.1}  & 96.9 \\ 
\hline
\end{tabular}}\vspace{-0.5cm}
\end{center}
\caption{Recognition rates of different methods using all bands on both datasets.}
\label{table:soa}
\end{table}

\paragraph*{\textbf{Comparison of HSI with RGB Image:}}
For this evaluation, we generate an RGB image for each of the hyperspectral image~\footnote{For transforming the HSI to RGB color space, we use (a) CIE 2006 tristimulus color matching functions, (b) CIE standard daylight illuminant ($D_{65}$), (c) and Silicon sensitivity of Hamamatsu camera}. In this regard, we refer the reader to the book by Ohta and Robertson~\cite{ohta} for detailed steps. Then, we apply the proposed method to an RGB image (with three channels)
in the same way as we applied to the hyperspectral image, discussed earlier in Section~\ref{sec:method}.

In Table \ref{table:soa1}, we  quantitatively evaluate the recognition accuracy of the whole band set in the hyperspectral cube, and compare it with a three channels RGB image. It is evident from the comparison that the classification performance for the computed features from HSI images  is significantly better than the computed features from RGB images on both datasets. In the literature, it has been shown by Pan et al.~\cite{spectralsignature}, NIR images exhibit a  distinct spectral properties of the skin and this information leads to more accurate classification in comparison to retained information in RGB images (i.e. visible range).  The reason being, in NIR range the spectral responses of the tissues are more discriminative due to larger penetration depth, which is dependent on the portion of melanin and hemoglobin~\cite{skin} in the human skin.

\begin{table}[t]

\begin{center}
\resizebox{6cm}{!} {
\begin{tabular}{|r| l l  l| }

\hline 
 	& HOG  &LBP  & DSIFT-FVs     \\
\hline
PolyU-HSFD$_{RGB}$ 		&85.0&78.8&85.4\\
PolyU-HSFD$_{All-Bands}$        &92.3&85.6&\textbf{96.1}\\
\hline
CMU-HSFD$_{RGB}$ 		&85.9&80.5&83.8	\\
CMU-HSFD$_{All-Bands}$ 		&91.5&86.1&\textbf{96.9}\\
\hline
\end{tabular}}\vspace{-0.5cm}
\end{center}
\caption{Comparison of hand-crafted HSI features with RGB image features for both datasets.}
\label{table:soa1}
\end{table}

\section{Conclusion} \label{sec:conclusion}

In this paper, we proposed  a novel pipeline for image-level  classification in the hyperspectral images. By doing this, we show that the discriminative spectral information at image-level features lead to significantly improved performance in a face recognition task. We also explored the potential of traditional feature descriptors in the hyperspectral images. From our evaluations, we observe that SIFT features outperform the  state-of-the-art hyperspectral face recognition methods, and  also the other descriptors.  With the increasing deployment of hyperspectral sensors in a multitude of applications, we believe that our approach  can effectively exploit the spectral information in hyperspectral images, thus beneficial to more accurate classification.

\newpage
{
\bibliographystyle{IEEEbib}
\bibliography{egbib}
}

\end{document}